\def\year{2018}\relax
\documentclass[letterpaper]{article} %DO NOT CHANGE THIS
\usepackage{aaai18}  %Required
\usepackage{times}  %Required
\usepackage{helvet}  %Required
\usepackage{courier}  %Required
\usepackage{url}  %Required
\usepackage{graphicx}  %Required
\usepackage{color}

\newcommand{\rpm}{\raisebox{.2ex}{$\scriptstyle\pm$}}

\begin{document}
\title{Towards Automated Let's Play Commentary}
\author{Matthew Guzdial, Shukan Shah, Mark Riedl\\
College of Computing\\
Georgia Institute of Technology\\
Atlanta, GA 30332\\
mguzdial3@gatech.edu, shukanshah@gatech.edu, riedl@cc.gatech.edu\\
}
\maketitle
\begin{abstract}

We introduce the problem of generating Let's Play-style commentary of gameplay video via machine learning. We propose an analysis of Let's Play commentary and a framework for building such a system. To test this framework we build an initial, naive implementation, which we use to interrogate the assumptions of the framework. We demonstrate promising results towards future Let's Play commentary generation.

\end{abstract}

\section{Introduction}

The rise of video streaming sites such as YouTube and Twitch has given rise to a new medium of entertainment, known as the ``Let's Play". Let's Plays involve video streamers providing commentary of their own gameplay to an audience of viewers. The impact of Let's Plays on today’s entertainment culture is evident from revenue numbers. Studies estimate that Let's Plays and other streamed game content will generate \$3.5 billion in ad revenue by the year 2021 \cite{juniperresearch2017}. What makes Let's Plays unique is the combination of commentary and gameplay, with each part influencing the other.

Let's Play serve two major purposes. First, to engage and entertain an audience through humor and interesting commentary. Second, to educate an audience, either explicitly as when a commentator describes what or why something happens in a game, or implicitly as viewers experience the game through the Let's Play. We contend that for these reasons, and due to the popularity and raw amount of existing Let's Play videos, this medium serves as an excellent training domain for automated commentary or explanation systems. 

Let's Plays could serve as a training base for an AI approach that learns to generate novel, improvisational, and entertaining content for games and video. 
Such a model could easily meet the demand for gameplay commentary with a steady throughput of new content generation. Beyond automating the creation of Let's Play we anticipate such systems could find success in other domains that require engagement and explanation, such as eSports commentary, game tutorial generation, and other domains at the intersection of education and entertainment in games. However, to the best of our knowledge, no existing attempt to automatically generate Let's Play commentary exists.

The remainder of this paper is organized as follows. First, we discuss Let's Plays as a genre and an initial qualitative analysis. Second, we propose a framework for the generation of Let's Plays through machine learning. Third, we cover relevant related work. Finally, we explore a few experimental, initial results in support of our framework.

\section{Let's Play Commentary Analysis}

For the purposes of this paper, we refer to the set of utterances made by players of Let's Plays as commentary. However, despite referring to these utterances as commentary, they are not strictly commenting or reflecting on the gameplay. This is in large part due to the necessity of a player to speak near constantly during a Let's Play video to keep the audience engaged. At times there is simply nothing occurring in the game to talk about. 

We analyzed hours of Let's Play footage of various Let's players and found four major types of Let's Play commentary. We list these types below in a rough ordering of frequency, and include a citation for a representative Let's Play video. However, we note that in most Let's Play videos, a player flows naturally between different types of commentary.

\begin{enumerate}
\item \textbf{Reaction: } The most common type of comment relates in some way to the gameplay occurring on screen \cite{reaction}. This can be descriptive, educational, or humorous. For example, reacting on a death by restating what occurred, explaining why it occurred, or downplaying the death with a joke.
\item \textbf{Storytelling: } The second most common comment we found was some form of storytelling that related events outside of the game. This storytelling could be biographical or fictional, improvised or pre-authored. For example, Hanson begins a fictional retelling of his life with ``at age six I was born without a face" in \cite{storytelling}.
\item \textbf{Roleplay: } Much less frequently than the first two types, some Let's players make comments in order to roleplay a specific character. At times this is the focus of an entire video and the player never breaks character, in others a player may slip in and out of the character throughout \cite{roleplay}. 
\item \textbf{ASMR: } ASMR stands for Autonomous Sensory Meridian Response \cite{barratt2015autonomous}, and there exists a genre of YouTube video dedicated to causing this response in viewers, typically for the purposes of relaxation. Let's players have taken notice, with some full Let's Plays in the genre of ASMR, and some Let's players slipping in and out of this type of commentary. These utterances tend to resemble whispering nonsense or making other non-word mouth noises \cite{asmr}. 

These high level types of commentary are roughly defined, and do not fully represent the variance of player utterances. These utterances also differ based on the number of Let's players in a single video, the potential for live interactions with an audience if a game is streamed, and variations among the games being played. We highlight these types as a means of demonstrating the breadth of distinct categories in Let's Play commentary.
Any artificial commentary generation system must be able to identify and generate within these distinct categories.

\end{enumerate}

\section{Proposed Framework}

In the prior section we list some high-level types of commentary we identified from Let's Play videos. This variance, in conjunction with the challenges present in all natural language generation tasks \cite{reiter2000building}, makes this problem an open challenge for machine learning approaches.

We propose the following two stage, high-level framework for machine learning approaches that generate Let's Play commentary. First, to handle the variance we anticipate the need for a pre-processing clustering step. This clustering step may require human action, for example, separating Let's Plays of a specific type, by Let's player, or by game. In addition or as an alternative, automated clustering may be applied to derive categories of utterances and associated gameplay footage. This may reflect the analysis we present in Section 2 or may find groupings specific to a particular dataset. 

In the second stage of our framework a machine learning approach is used to approximate a commentary generation function. The most naive interpretation of this would be to learn a mapping between gameplay footage and commentary based on the clustered dataset from the first step. However, we anticipate the need to include prior commentary and its relevant gameplay footage as input to generate coherent commentary. We expect a full-fledged implementation of this approach would make use of state of the art natural language generation methods.

\begin{figure*}[tb]
\centering
	\includegraphics[width=\linewidth]{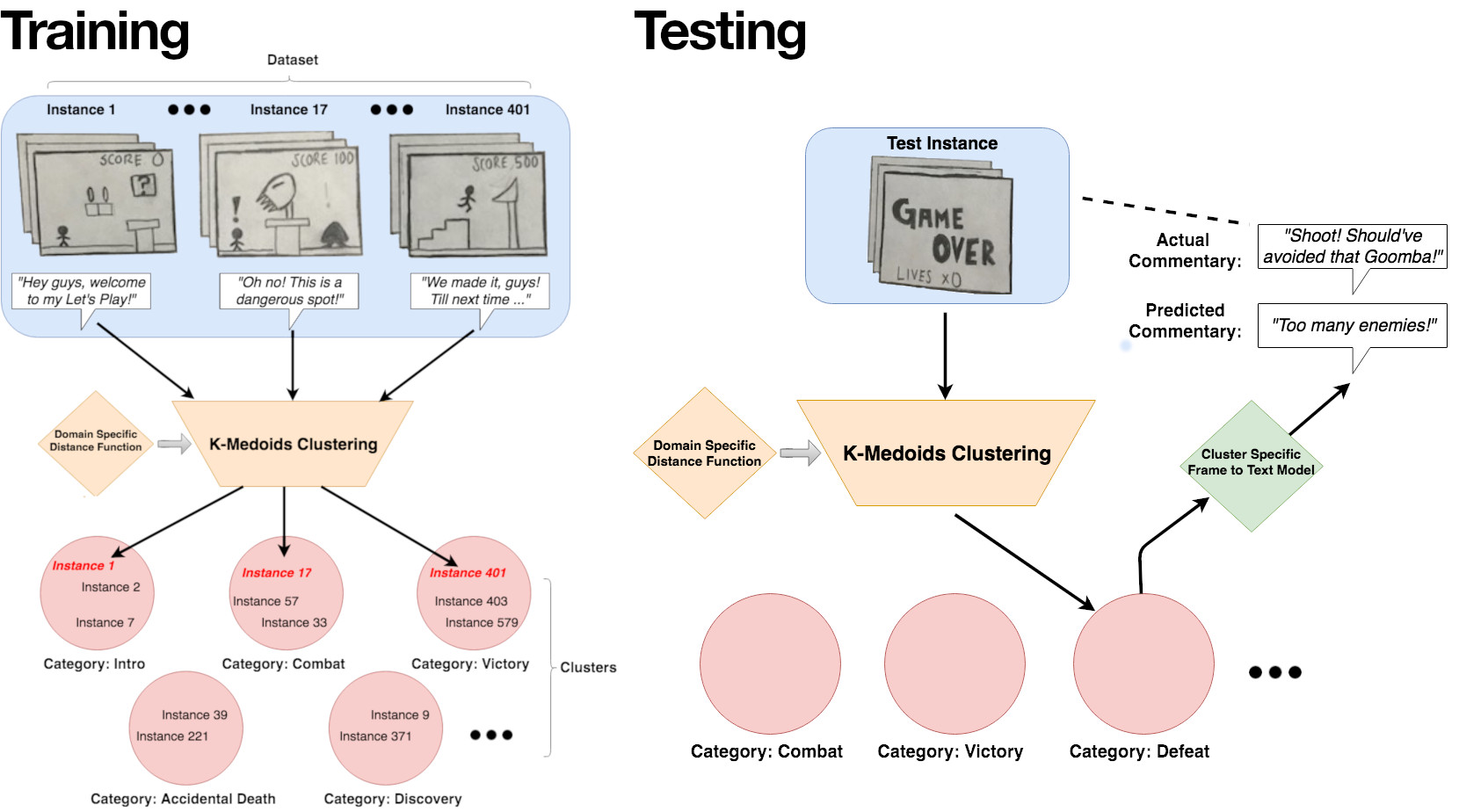}
	\caption{Visual representation of our experimental implementation.}
	\label{fig:systemOverview}
\end{figure*}

\section{Related Work}

There exists prior work on automatically generating textual descriptions of gameplay. Bardic \shortcite{barot2017bardic} creates narrative reports from Defense of the Ancients 2 (DOTA 2) game logs for improved and automated player insights. This has some similarity to common approaches for automated journalism of physical sports \cite{graefe2016guide} or automated highlight generation for physical sports \cite{kolekar2006event}. These approaches require a log of game events. Our proposal is for live commentary of an ongoing video game stream.
Harrison et al. \shortcite{rational2017harrison} create explanations of an AI player's actions for the game Frogger. All of these approaches depend on access to a game's engine or the existence of a publicly accessible logging system.

To the best of our knowledge, there have been no AI systems that attempt to directly generate streaming commentary of live gameplay footage. Nonetheless, work exists that maps visual elements (such as videos and photos) to story-like natural language. For example, The Sports Commentary Recommendation System or SCoReS \shortcite{scores2017lee} and PlotShot \shortcite{plotshot2016cardona} seek to create stories centered around visual elements measured/recorded by the system. While SCoReS learns a mapping from specific game states to an appropriate story \cite{scores2017lee}, PlotShot is a narrative planning system that measures the distance between a photo and the action it portrays \cite{plotshot2016cardona}. 
Our domain is different as we are not illustrating stories but rather trying to construct live commentary on the fly as the system receives a continuous stream of events as input.

Significant prior work has explored Let's Play as cultural artifact and as a medium. For example, prior studies of the audience of Let's Plays \cite{sjoblom2017people}, content of Let's Plays \cite{sjoblom2017content}, and building communities around Let's Play \cite{hamilton2014streaming}. The work described in this paper is preliminary as a means of exploring the possibility for automated generation of Let's Play commentary. We anticipate future developments in this work to more closely engage with scholarship in these areas.

More recent work explores the automated generation of content from Let's Plays, but not automated commentary. Both Guzdial and Riedl \shortcite{guzdial2016game} and Summerville et al. \shortcite{summerville2016learning} look to use Longplay's, a variation of Let's Play generally without commentary, as part of a process to generate video game levels through procedural content generation via machine learning \cite{summerville2017procedural}. Other work has looked at eSport commentators in a similar manner, as a means of determining what approaches the commentators use that may apply to explainable AI systems \cite{dodge2018experts}. However, this work only presented an analysis of this commentary, not of the video, and without any suggested approach to generate new commentary.

\section{Experimental Implementation}

In this section we discuss an experimental implementation of our framework in the domain of Super Mario Bros. Let's Plays. The purpose of this experimental implementation is to allow us to interrogate the assumptions implicit in our framework, in particular that clustering is a necessary and helpful means of handling the variance of Let's Play commentary.

We visualize a high-level overview of our implementation in Figure \ref{fig:systemOverview}. As a preprocessing step we take video and an associated automated transcript from YouTube, to which we apply ffmpeg to break the video into individual frames and associated commentary lines or utterances. 
We make use of 1 FPS when extracting frames from video, given most comments took at least a few seconds. Therefore most comments are paired with two or more frames in a sequence, representing the video footage while the player makes that comment.

A dataset of paired frames and an associated utterance serves as input into our training process. We represent each frame as a ``bag of sprites", based on the bag of words representation used in natural language processing. We pull the values of individual sprites from each frame using a spritesheet of Super Mario Bros. with the approach described in \cite{guzdial2016game}. We combine multiple bags of sprites when a comment is associated with multiple frames. We also represent the utterance as a bag of words as well. 
In this representation we cluster the bag of sprites and words according to a given distance function with K-Medoids. We determine a value of $k$ for our medoids clustering method according to the distortion ratio \cite{pham2005selection}. This represents the first step of our framework.

For the second step of our framework in this implementation, we approximate a frame-to-commentary function using one of three variations. These variations are as follows: 

\begin{itemize}
  \item \textbf{Random: } Random takes the test input, and simply returned a random training element's comment data.
  \item \textbf{Forest: } We constructed a 10 tree random forest, using the default SciPy random forest implementation \cite{jones2014scipy}. We used all default parameters except that we limited the tree depth to 200 to incentive generality. This random forest predicted from the bag of sprites frames representation to a bag of words comment representation. This means it produced a bag of words instead of the ordered words necessary for a comment, but it was sufficient for comparison with true comments in the same representation. 
\item \textbf{KNN: } We constructed two different $K$-Nearest Neighbor (KNN) approaches, based on a value of $k$ of 5 or 10. In this approach we grabbed the $k$ closest training elements to a test element according to frame distance. Given these $k$ training examples, we took the set of their words as the output. As with the Forest baseline, this does not represent a final comment. We note further that there are far too many words for a single comment using this method, which makes it difficult to compare across the other baselines, but we can compare between variations of this baseline. 
\end{itemize}

For testing we then take as input some frames without or with withheld commentary, determine which cluster it would be clustered into, and use a learned frame-to-text function to predict an output utterance. This can then be compared to the true utterance if there is one. We note this is the most naive possible implementation, and true attempts at this problem will want to include as input to this function prior frames and prior generated comments, in order to determine an appropriate next comment.

Clustering approaches require a distance function to determine the distance between any two arbitrary elements you might wish to cluster. Our distance function is made up of two component distance functions; one part measures the distance between the frame data (represented as a bag of sprites) and one part measures the distance between the utterances (represented as a bag of words).
For both parts we make use of cosine similarity.
Cosine similarity, a simple technique used popularly in statistical natural language processing, measures how similar two vectors are by measuring the angle between them. The smaller the angle, the more similar the two vectors are (and vice versa). An underlying assumption that we make is that the similarity between a pair of comments is more indicative of closeness than similar frame data, simply because it is very common for two instances to share similar sprites (especially when they're adjacent in the video). Thus, when calculating distance, the utterance cosine similarity is weighted more heavily (75\%) than the frame cosine similarity (25\%).

\begin{figure}[tb]
	\includegraphics[width=\columnwidth]{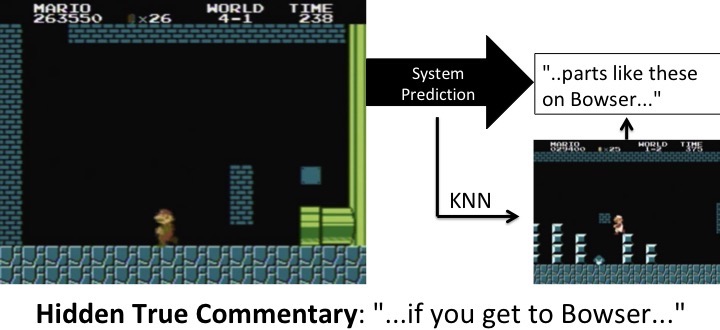}
	\caption{An example test commentary prediction. The true frame on the left is associated with the utterance ``if you get to Bowser" and our method returns "parts like these on Bowser".}
	\label{fig:example}
\end{figure}

At present, we use Super Mario Bros. (SMB),  a well-known and studied game for the NES, as the domain for our work. We chose Super Mario Bros because prior work in this domain has demonstrated the ability to apply machine learning techniques to scene understanding of SMB gameplay \cite{guzdial2016game}. 

For our initial experiments we collected a dataset of two fifteen minute segments of a popular Let's Player playing through Super Mario Bros. along with the associated text transcripts generated by Youtube. We use one of these videos as a training set and one as a test set. This means roughly three-hundred frame and text comment pairs in both the training and testing sets, with 333 for the training set and 306 for the testing set. This may seem small given that each segment was comprised of fifteen minutes of gameplay footage, however it was due to the length and infrequency of the comments. 

We applied the training dataset to our model. In clustering we found $k=6$ for our K-medoids clustering approach according to the distortion ratio. This lead to six final clusters. These clusters had a reasonable spread, with a minimum of 31 elements, a maximum of 90 elements, and a median size of 50. From this point we ran our two evaluations. 

\begin{table}[tb]
\caption{Comparison of different ML approaches to the frame-to-comment function approximation, trained across all available training data or only within individual clusters. Values are average cosine distance, thus lower is better.}
\label{tab:variationsResults}
\begin{center}
\begin{tabular}{ |l|c|c| } 
 \hline
  & Standard & Per-Cluster \\ 
  \hline
 Random & 0.940\rpm0.102 & 0.937\rpm0.099\\ 
  \hline
 Forest & 0.993\rpm0.042 & 0.970\rpm0.090\\ 
  \hline
 KNN 5 & 0.886\rpm0.091 & 0.885\rpm0.100\\ 
  \hline
 KNN 10 & 0.854\rpm0.091 & 0.852\rpm0.101\\ 
  \hline
\end{tabular}
\end{center}
\end{table}

\subsection{Standard vs. Per-Cluster Experiment} 

For this first experiment we wished to interrogate our assumption that automated clustering represented a cost-effective means of handling the variance of Let's Play commentary. To accomplish this, we trained each of our three variations (Random, Forest, KNN) according to two different processes. In what we call the standard approach each variation is trained on the entirety of the dataset. In the per-cluster approach a model of each variation is trained for each cluster. This means that we had one random forest in the standard approach, and six random forests (one for each cluster) in the per-cluster approach.

We tested all 306 test elements for each approach. For the per-cluster approaches we first clustered each test element only in terms of its frame data, and then used the associated model trained only on that cluster to predict output. In the standard variation we simply ran the test element's frame data through the trained function. For both approaches we compare the cosine distance of the true withheld comment and the predicted comment. This can be understood as the test error of each approach, meaning a lower value is better. If the per-cluster approach of each variation outperforms the standard approach, then that would be evidence that the smaller training dataset size of the per-cluster approach was more than made up for by the reduction in variance.

We compile the results of this first experiment in Table \ref{tab:variationsResults}. In particular, we give the average cosine distance between the predicted and true comment and the standard deviation. In all cases the per-cluster variation outperforms the standard variation. The largest improvement came from the Random Forest baseline, while the KNN baseline with $k=5$ had the smallest improvement. This makes sense as even in the standard approach, KNN would largely find the same training examples as the per-cluster approach. However, the actual numbers do not matter here. Given the size of the associated datasets, seeing any trend indicates support for our assumption, which we would anticipate to scale with larger datasets and more complex methods.

\begin{table}[tb]
\caption{Comparison of different ML approaches to the frame-to-comment function approximation, trained with the true cluster or a random cluster. Values are average cosine distance, thus lower is better.}
\label{tab:randomResults}
\begin{center}
\begin{tabular}{ |l|c|c| } 
 \hline
  & True Cluster & Random Cluster \\ 
  \hline
 Random & 0.937\rpm0.099 & 0.936\rpm0.101\\ 
  \hline
 Forest & 0.970\rpm0.090 & 0.986\rpm0.058\\ 
  \hline
 KNN 5 & 0.885\rpm0.100 & 0.901\rpm0.077\\ 
  \hline
 KNN 10 & 0.852\rpm0.101 & 0.885\rpm0.066\\ 
  \hline
\end{tabular}
\end{center}
\end{table}

\subsection{Cluster Choice Experiment}

The prior experiment suggests that our clusters have successfully cut down on the variance of the problem. However, this does not necessarily mean that our clusters represent any meaningful categories of or relationships between frames and comments. Instead, the performance from the prior experiment may be due to the clusters reducing the dimensionality of the problem. It is well-recognized that even arbitrary reductions in the dimensionality of a problem can lead to improved performance for machine learning approaches \cite{bingham2001random}. This interpretation would explain the improvement seen in our random forest baseline, given that this method can be understood as including a feature selection step.  Therefore we ran a secondary experiment in which we compared the per-cluster variations using either the assigned cluster or a random cluster. If it is the case that the major benefit to the approaches came from the dimensionality reduction from the cluster, we would anticipate equivalent performance no matter which cluster is chosen. 

We summarize the results of this experiment in Table \ref{tab:randomResults}. Outside of the random variation, the true cluster faired better than a random cluster. In the case of the random baseline the performance was marginally better with a random cluster, but nearly equivalent. This makes sense given that the random variation only involves uniformly sampling across the cluster distribution as opposed to learning some mapping from frame representation to comment representation. 

These results indicate that the clusters do actually represent meaningful types of relationships between frame and text. This is further evidenced looking at the average comment cosine distance between the test examples and the closest medoid according to the test example's frame (0.910\rpm0.115) and a randomly selected medoid (0.915\rpm0.108). 

\subsection{Qualitative Example}

We include an example of output in Figure \ref{fig:example} using a KNN with $k=1$ in order to get a text comment output as opposed to a bag of words. Despite the text having almost nothing to do with the frame in question the text is fairly similar from our perspective. It is worth noting again that all our data comes from the same Let's Player, and therefore may represent some style of that commentator.

\section{Limitations and Future Work}

In this paper we introduce the problem of creating a machine learning approach for Let's Play commentary generation. Towards this purpose we present an abstract framework for solving this problem and present a limited, experimental implementation which we interrogate. We find some results that present initial evidence towards assumptions in our framework. However, due to the scale of this experiment (two Let's Play videos for one game), we cannot state with certainty that these results will generalize.

Future implementations of this framework will incorporate more sophisticated methods for natural language processing to generate novel commentary. As stated above, for the purposes of this initial experiment we went with the naive approach of predicting output commentary solely from an associated sequence of gameplay video frames. We anticipate a more final system will require a history of prior comments and associated frames.

In this initial work we drew upon random forest, and KNN as a means of generating output commentary, represented as a bag of words. We note that as our training data increased, we would anticipate a runtime increase for both these approaches (though we can limit this in the random forest by limiting depth). If we want to have commentary generated in real-time, we might instead want to make use of a deep neural network or similar model of fixed size.

Besides generating entertaining commentary for Let's Plays, a final working system could be useful in a variety of settings. One obvious approach would be to attempt to extend such a system to color commentary for eSports games \cite{dodge2018experts}. More generally, such a system might help increase user engagement with AI agents by aiding in Explainable AI approaches to rationalize decision-making \cite{rational2017harrison}. 

\section{Conclusions}

In this paper, we define the problem of automatic commentary of Let's Play videos via machine learning. Our framework requires an initialize clustering stage to cut back on the implicit variance of Let's Play commentary, followed by a function approximation for commentary generation. We present an experimental implementation and multiple experimental results. Our results lend support to our framework. Although there is much to improve upon, the work is an exciting first step towards solving the difficult problem of automated, real-time commentary generation.

\section{Acknowledgements}

This material is based upon work supported by the National Science Foundation under Grant No. IIS-1525967.

\bibliographystyle{aaai}
\bibliography{exag18}

\end{document}